\documentclass[runningheads]{llncs}
\usepackage{graphicx}
\usepackage{amssymb}
\begin{document}
\title{Tracking of Micro Unmanned Aerial Vehicles: A Comparative Study}
%
%
\author{Fatih G\"{O}K\c{C}E\orcidID{0000-0001-7935-7982}}
%
%
\institute{Department of Computer Engineering\\ S\"{u}leyman Demirel University, 32260, Isparta, TURKEY \\
	\email{fatihgokce@sdu.edu.tr}}
\maketitle              
\begin{abstract}
Micro unmanned aerial vehicles (mUAV) became very common in recent years. As a result of their widespread usage, when they are flown by hobbyists illegally, crucial risks are imposed and such mUAVs need to be sensed by security systems. Furthermore, the sensing of mUAVs are essential for also swarm robotics research where the individuals in a flock of robots require systems to sense and localize each other for coordinated operation. In order to obtain such systems, there are studies to detect mUAVs utilizing different sensing mediums, such as vision, infrared and sound signals, and small-scale radars. However, there are still challenges that awaits to be handled in this field such as integrating tracking approaches to the vision-based detection systems to enhance accuracy and computational complexity. For this reason, in this study, we combine various tracking approaches to a vision-based mUAV detection system available in the literature, in order to evaluate different tracking approaches in terms of accuracy and as well as investigate the effect of such integration to the computational cost.

\end{abstract}

\section{Introduction}
Widespread availability of micro unmanned aerial vehicles (mUAV) which arises due to technological advancement in micro-controller and sensor technologies, and lowering costs in recent decades imposed a common thread where intruder mUAVs are required to be sensed by early warning systems. Besides, the sensing of mUAVs is a also a compelling problem for swarm robotics research where multiple agents cooperate to accomplish a common task. Sensing and localizing other agents in a swarm is a critical problem for the success of the task.

In order to sense mUAVs, various approaches are utilized in the literature, such as vision~\cite{Lin2014,Zhang2014,Lai2011}, infrared~\cite{roberts1,roberts2,stirling,Welsby01,Raharijaona2015,etter}, sound signals~\cite{epflAudio,Tijs2010,Nishitani06,maxim08,rivard08}, small-scale radars~\cite{Moses2011,Moses2014}, and the Global Positioning System (GPS)~\cite{vasarhelyi2014outdoor,Brewer14}. However, each	of these approaches put their own limitations regarding the sensing distance, and also the scenario in which they are applicable. For example, a system developed for swarm robotics purposes may not be applicable to sensing an intruder mUAV if specific equipments are put on the platforms in the swarm.  

In the current study, we aim to sense mUAVs using only visual appearance cues since this modality is better suited for both intruder mUAV sensing and swarm robotics. In this regard, our problem fits into object recognition problem widely studied in computer vision and pattern recognition~\cite{andreopoulos201350}. However, when mUAVs are considered as the specific object to be recognized, this area of research is relatively new. 

Sensing mUAVS in video frames is a challenging problem due to some inherent aspects of the problem: (a)~Non-convex structure of the mUAVs causes the bounding box to include various complex and challenging background patterns. (b)~Rotation and tilt of mUAVs and also illumination changes create very different appearances of the same mUAV. (c)~When the mUAV or the camera is moving, motion blur can occur in the images. (d)~In swarm robotics applications a real-time solution is required with the limited processing power on the platforms. 

In the literature, there are studies proposing methods to detect mUAVs in video frames~\cite{Lin2014,Zhang2014,Lai2011,Opromolla18,Gokce15}. However, detection only methods suffer from high computational complexity, since they build a large model and test is on the current frame to determine the presence and bounding box of the mUAV. At this point, when we consider to speed up the sensing process of mUAVs, we can integrate a tracking method since tracking is faster than detection as a relatively lightweight model is trained using an initial frame. Tracking can also help to increase sensing performance by giving estimations for the frames where the detector is not able to output a bounding box estimate.  

Various trackers are proposed in the literature with different approaches~\cite{Grabner06,Babenko11,Henriques15,Kalal12,Kalal10,Held16,Bolme10,Lukezic18}. When it comes to integrate a detector with a tracker, we need an investigation of different methods to decide on which one to choose among various options. To the best of our knowledge there is no such investigation comparing tracking methods for mUAV tracking. There are only a few studies considering to combine detection and tracking~\cite{Lin2014,Opromolla18,Aker2018} without any comparison. In~\cite{Lin2014}, Kalman Filter is utilized for tracking. A template matching based tracking method is proposed in~\cite{Opromolla18}. In~\cite{Aker2018}, a Neural Turing Machine based tracker model named as Neural Turing Tracker is proposed. Therefore, in order to fill this gap, in this study, we present a comprehensive comparison among various trackers investigating their effects on sensing performance and as well as computational complexity on specifically mUAV tracking problem.

The rest of the paper is organized as follows. The next section describes our methodology and introduces the tracker methods involved. Section~\ref{datasetMetrics} presents our dataset and defines the metrics we utilized in our evaluations. In Section~\ref{results}, we provide our experimental results. Section~\ref{conclusions} concludes the paper.

\section{Methodology}\label{sec:methodology}

In order to compare the trackers, we employed the approach known as the track-by-detection in the literature~\cite{Luo2018}. This approach requires a detector and tracker to work in conjunction. Since our aim is to compare various trackers, we selected the mUAV detector proposed in~\cite{Gokce15} as our baseline. This detector, called as \mbox{C-LBP}, is based on boosted cascaded classifiers and uses local binary bit patterns as feature descriptor. It is reported to perform better than other two methods based on HAAR features and histogram of gradients in terms of both detection performance and speed. 

\begin{figure}[]
	\begin{center}
		\includegraphics[width=\textwidth]{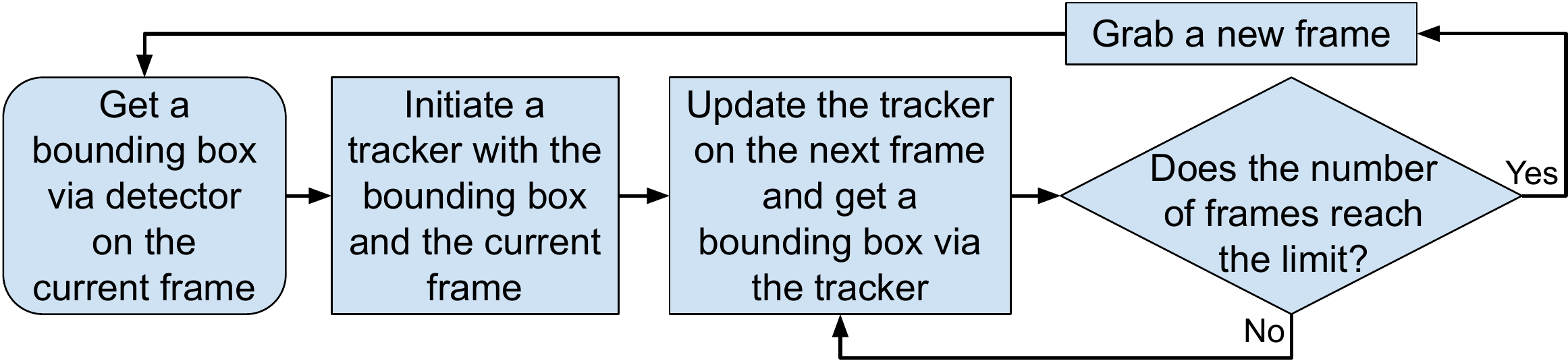}
		\caption{Block diagram for the track-by-detection approach.}
		\label{fig:Methodology}
	\end{center}
\end{figure}

Figure~\ref{fig:Methodology} illustrates the track-by-detection applied in this study. We get a bounding box via the detector on the very first frame; then, we initialize a tracker and use it to estimate the bounding box of the mUAV on the following frames. However, since the trackers are prone to drifting~\cite{Luo2018}, the detector still has to be re-utilized after some number of frames to initialize the tracker again. This number of frames plays a critical role affecting the balance between the computational time and the performance of the system, and needs investigation. For this reason, we will name this number as the frame limit (\textit{f-lim}), and evaluate the trackers by varying the \textit{f-lim} value. We should note that, if the tracker fails to give an estimate on a frame, we continue by applying the tracker to the next frame without running the detector on the failed frame until reaching the \textit{f-lim} value, since we want to understand the effect of a certain \textit{f-lim} value in each test. In addition, if the detector outputs more than one bounding boxes when it is utilized, we ignore these detections and keep on using previously initialized tracker. Assuming that we know only one quadrotor in a frame, we regarded multiple bounding boxes returned from the detector as a detection failure. The detector is able to detect the quadrotor on the first frames of the test videos. 

As the trackers require less computational power, the track-by-detection method results in a speed-up. Moreover, if the trackers can give estimated bounding boxes for the frames where the detector would fail, the performance when compared with the detection only methods can be improved due to reduction in the number of false negatives. 

We selected $8$ different trackers to be compared in this study. Table~\ref{tab:tableTrackers} presents these trackers along with their important aspects. The trackers in this table are ordered regarding their proposal time. We utilized OpenCV~\cite{opencv} implementations of these tracking methods.

\begin{table}[]
	\centering
	\caption{Trackers compared in this study are listed along with their important aspects and corresponding references.}
	\label{tab:tableTrackers}
	\resizebox{\textwidth}{!}{%
		\begin{tabular}{|l|l|c|}
			\hline
			\textbf{Tracker} & \textbf{Important aspects}                                                                                                                                                                                                                                                                                                                                                                                                      & \textbf{Reference} \\ \hline
			BOOSTING         & \begin{tabular}[c]{@{}l@{}}Tracking method based on online AdaBoost feature selection. \\ The method utilizes surrounding background patterns \\ as negative samples in update phase to overcome drifting.\end{tabular}                                                                                                                                                                                                         & \cite{Grabner06}   \\ \hline
			MOSSE            & \begin{tabular}[c]{@{}l@{}}Study proposing correlation filters for visual tracking for \\ the first time through Minimum Output Sum of Squared \\ Error (MOSSE) filter initialized using a single frame and\\ computing the correlation in the Fourier domain.\end{tabular}                                                                                                                                                     & \cite{Bolme10}     \\ \hline
			MEDIANFLOW       & \begin{tabular}[c]{@{}l@{}}Lucas-Kanade tracker based tracker where the objects are \\ tracked in a forward and backward (FB) fashion in time and \\ the error between FB trajectories are minimized to compute \\ the estimated trajectory of the object.\end{tabular}                                                                                                                                                         & \cite{Kalal10}     \\ \hline
			MIL              & \begin{tabular}[c]{@{}l@{}}Extension of BOOSTING method via multiple instance \\ learning where the tracker is updated using not only \\ the current positive patch but also using its neighbouring patches.\end{tabular}                                                                                                                                                                                                    & \cite{Babenko11}   \\ \hline
			TLD              & \begin{tabular}[c]{@{}l@{}}Standing for tracking, learning and detection, this method takes \\ long-term tracking problem as a combination of separate \\ simultaneous tracking, learning and detection tasks. MEDIANFLOW\\ is utilized for tracking. PN learning paradigm is proposed and \\ utilized to augment the training set of the detector in every frame \\ so that the detector is continously improved.\end{tabular} & \cite{Kalal12}     \\ \hline
			KCF              & \begin{tabular}[c]{@{}l@{}}A correlation filter based approach where circulant property \\ of translated image patches is exploited in the Fourier domain \\ to train the tracker efficiently.\end{tabular}                                                                                                                                                                                                                     & \cite{Henriques15} \\ \hline
			GOTURN           & \begin{tabular}[c]{@{}l@{}}A deep learning based tracking approach where an offline \\ model is trained beforehand and utilized for tracking without \\ requiring online training.\end{tabular}                                                                                                                                                                                                                                 & \cite{Held16}      \\ \hline
			CSRT  & \begin{tabular}[c]{@{}l@{}} Discriminative correlation filter based method which constructs\\ a spatial reliability map for adapting the filter support to the portion\\ of the selected area from the frame for tracking. In this way,\\ problems due to circular shift and rectangular shape assumption\\ are overcame.\end{tabular}
			& \cite{Lukezic18}   \\ \hline
		\end{tabular}%
	}
\end{table}

\section{Dataset and Evaluation Metrics}\label{datasetMetrics}

We utilized the dataset proposed in~\cite{Gokce15}. Although the complete dataset includes both indoor and outdoor data, we only used three videos from its outdoor part. Each of these three videos have different characteristics. In each video, a quadrotor is flown in front of a fixed camera outdoors. In the first video, the quadrotor is flown in a calm manner while in the second one, it is flown with more agility. In both of these two videos, the background is stationary. In the last video, however, there are moving background objects like motorcycles, pedestrians, cars and buses. The flight maneuvers in the third video are calm. These three videos will be called as (i)~calm, (ii)~agile, and (iii)~moving background in the remaining parts of the paper. These videos have 2954, 3823, and 3900 frames, respectively, with a resolution of $1280\times720$. 

In order to evaluate the accuracy of tracking, we utilized F-Score metric which is defined as:
\begin{equation}\label{fscore}
	F\textrm{\textit{-}}Score=2\times\frac{Precision\times Recall}{Precision+Recall}.
\end{equation}
Here, precision is defined as: 
\begin{equation}\label{precision}
	Precision=\frac{tp}{tp+fp},
\end{equation}
where $tp$ and $fp$ are the numbers of true positives and false positives, respectively. Recall is defined as:
\begin{equation}\label{recall}
	Recall=\frac{tp}{tp+fn},
\end{equation}
where $fn$ corresponds to the number of false negatives. 

A bounding box obtained from the detector or tracker ($B_o$) is counted as a true positive if its Jaccard index (J)~\cite{Jaccard} is greater than $60\%$. J is calculated as follows:
\begin{equation}\label{jaccard}
	J(B_o,B_g)=\frac{|B_o\cap B_g|}{|B_o\cup B_g|},
\end{equation}
where $B_g$ denotes the ground truth bounding box. If J is not greater than $60\%$, $B_o$ is considered as a $f_p$. If no $B_o$ is reported for an image frame, then $fn$ is increased by one.

For evaluating the computational time, we utilized three different times, namely, (i)~initiation time of the tracker (including the detection time), (ii)~update time of the tracker and (iii)~average required time calculated as the total time required to process a video (including all initiation and update times) divided by the number of frames in the corresponding video.

\section{Results}\label{results}

In order to compare the trackers, we have evaluated them in terms of both tracking accuracy and computational time. In accuracy evaluation experiments, we tested each tracking method employing the track-by-detection framework presented in Section~\ref{sec:methodology} on each of the videos. For each of the tracking methods, we also varied \textit{f-lim} value between $10$ and $100$ with the increments of $10$ to understand its effect.

\begin{figure}[]
	\begin{center}
		\includegraphics[width=0.8\textwidth]{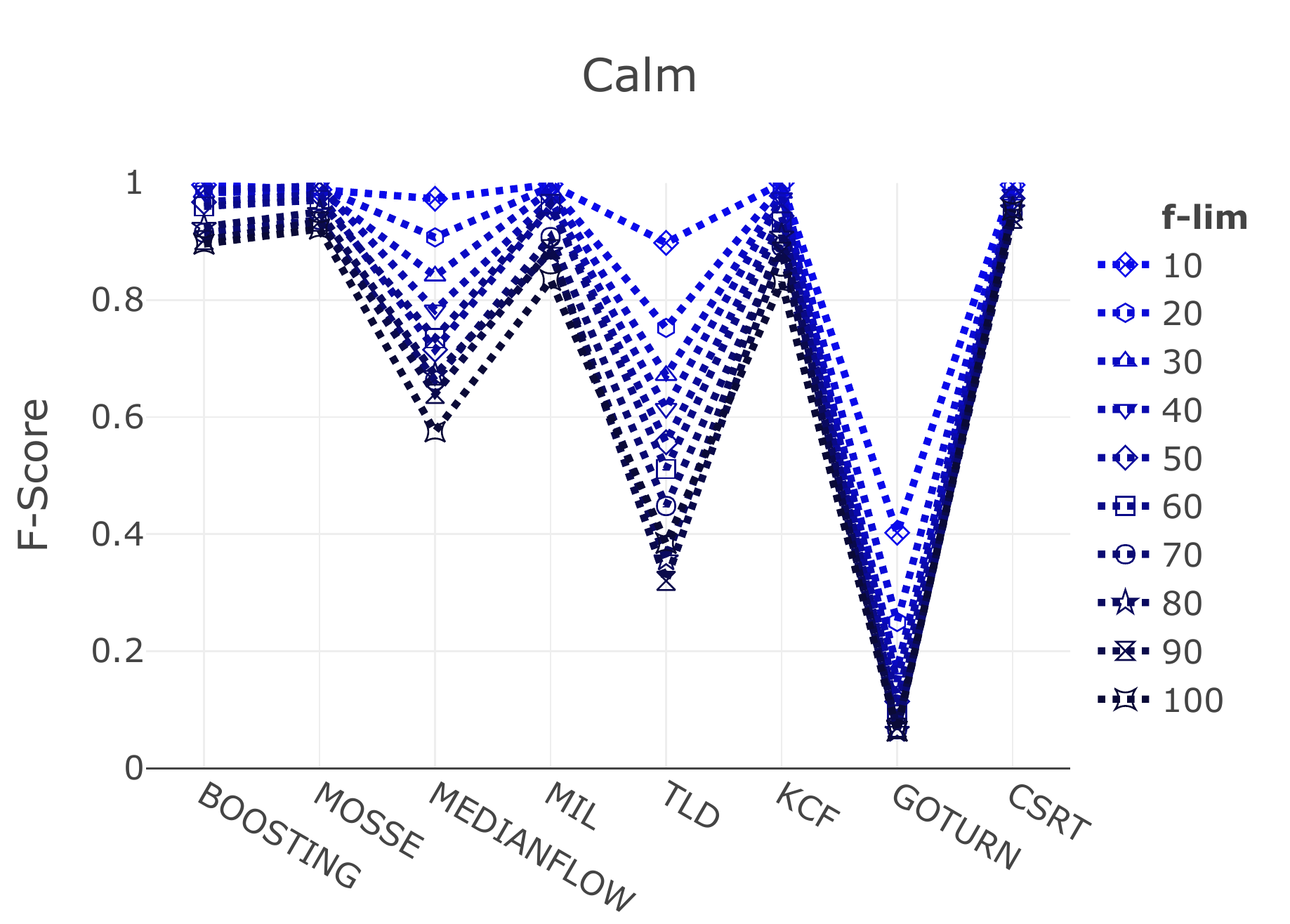}
		\caption{F-Score curves for \textit{calm} video. Higher values of F-Score indicate better performance.}
		\label{fig:FScoreCalm}
	\end{center}
\end{figure}

\begin{figure}[]
	\begin{center}
		\includegraphics[width=0.8\textwidth]{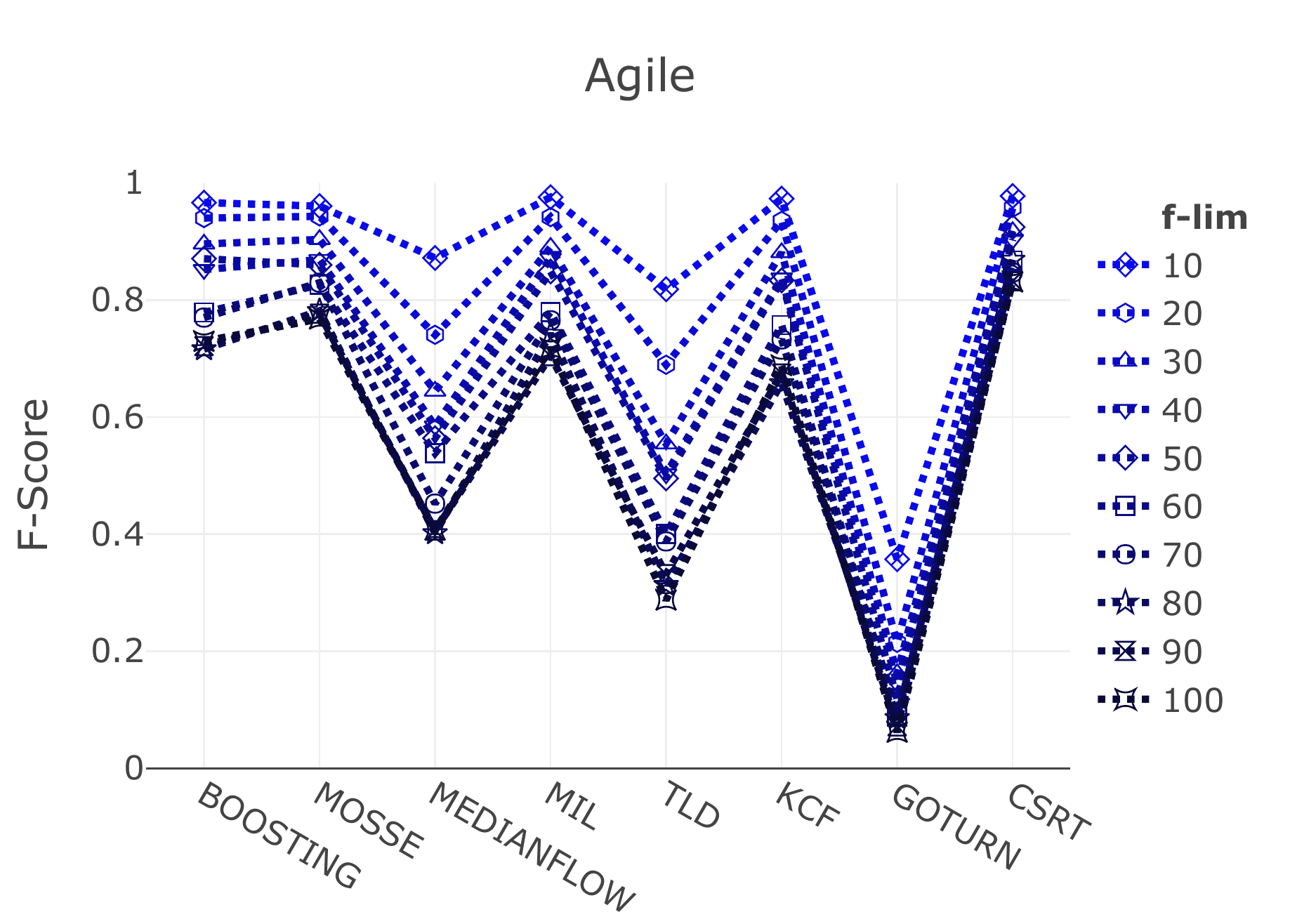}
		\caption{F-Score curves for \textit{agile} video.}
		\label{fig:FScoreAgile}
	\end{center}
\end{figure}

\begin{figure}[]
	\begin{center}
		\includegraphics[width=0.8\textwidth]{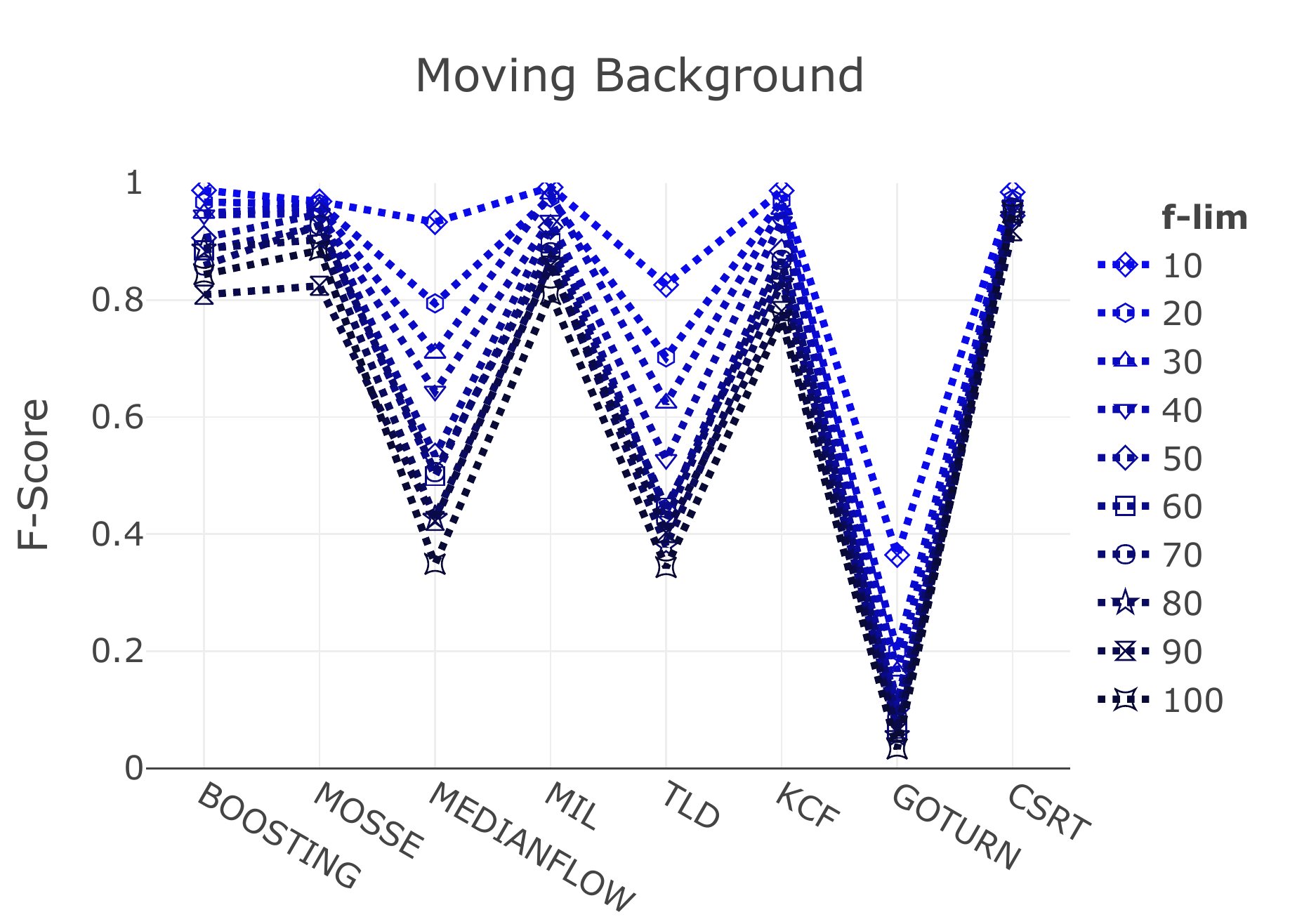}
		\caption{F-Score curves for \textit{moving background} video.}
		\label{fig:FScoreMovingbg}
	\end{center}
\end{figure}

\begin{figure}[]
	\begin{center}
		\includegraphics[width=0.8\textwidth]{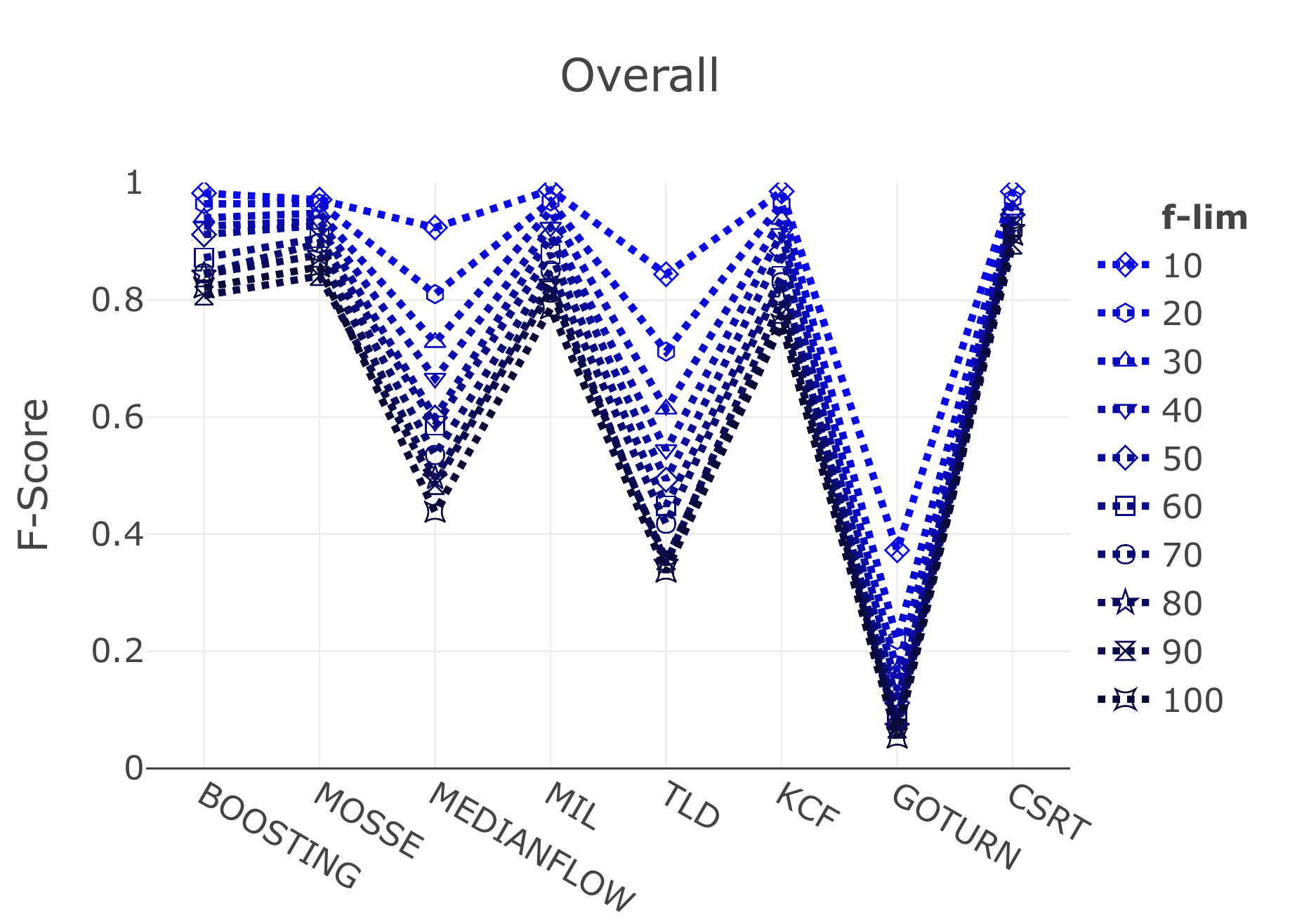}
		\caption{F-Score curves for all videos combined.}
		\label{fig:FScoreOverall}
	\end{center}
\end{figure}

Figure~\ref{fig:FScoreCalm},~\ref{fig:FScoreAgile}, and~\ref{fig:FScoreMovingbg} illustrate the results of the accuracy evaluation experiments, respectively, for calm, agile, and moving background videos. In addition, overall results for these three videos are given in Figure~\ref{fig:FScoreOverall}. 

When Figure~\ref{fig:FScoreCalm} and~\ref{fig:FScoreAgile} are compared, we can deduce that agile maneuvers influence all the trackers negatively, since F-Scores of all trackers are getting lower values for the agile video. When we look at the range and distribution of F-Score values in general for different values of \textit{f-lim}, CSRT is outperforming the others, then MOSSE, BOOSTING, MIL and KCF are coming in order. However, for lower values of \textit{f-lim}, KCF and MIL are performing better or at least comparable when compared to MOSSE and BOOSTING.

If we compare Figure~\ref{fig:FScoreCalm} and~\ref{fig:FScoreMovingbg}, the performances of all the trackers are decreasing when there are moving objects in the background. Comparing Figure~\ref{fig:FScoreAgile} and~\ref{fig:FScoreMovingbg} for CSRT, MOSSE, BOOSTING, MIL and KCF, we infer that these trackers are more prone to the agility than the moving background objects.

Overall performances depicted in Figure~\ref{fig:FScoreOverall} indicate that CSRT, MOSSE, BOOSTING, MIL and KCF are the best performing $5$ methods in order. BOOSTING is better than MOSSE only for $\textit{f-lim}=20$. When F-Score and \textit{f-lim} values are inspected together for \textit{f-lim} values of $10$ and $20$, BOOSTING, MOSSE, MIL, KCF and CSRT methods are resulting in a better performance than the baseline detection only approach with F-Score values always above $0.957$ which is reported as the overall F-Score value of C-LBP detection method in~\cite{Gokce15}. 

\begin{figure}[]
	\begin{center}
		\includegraphics[width=0.7\textwidth]{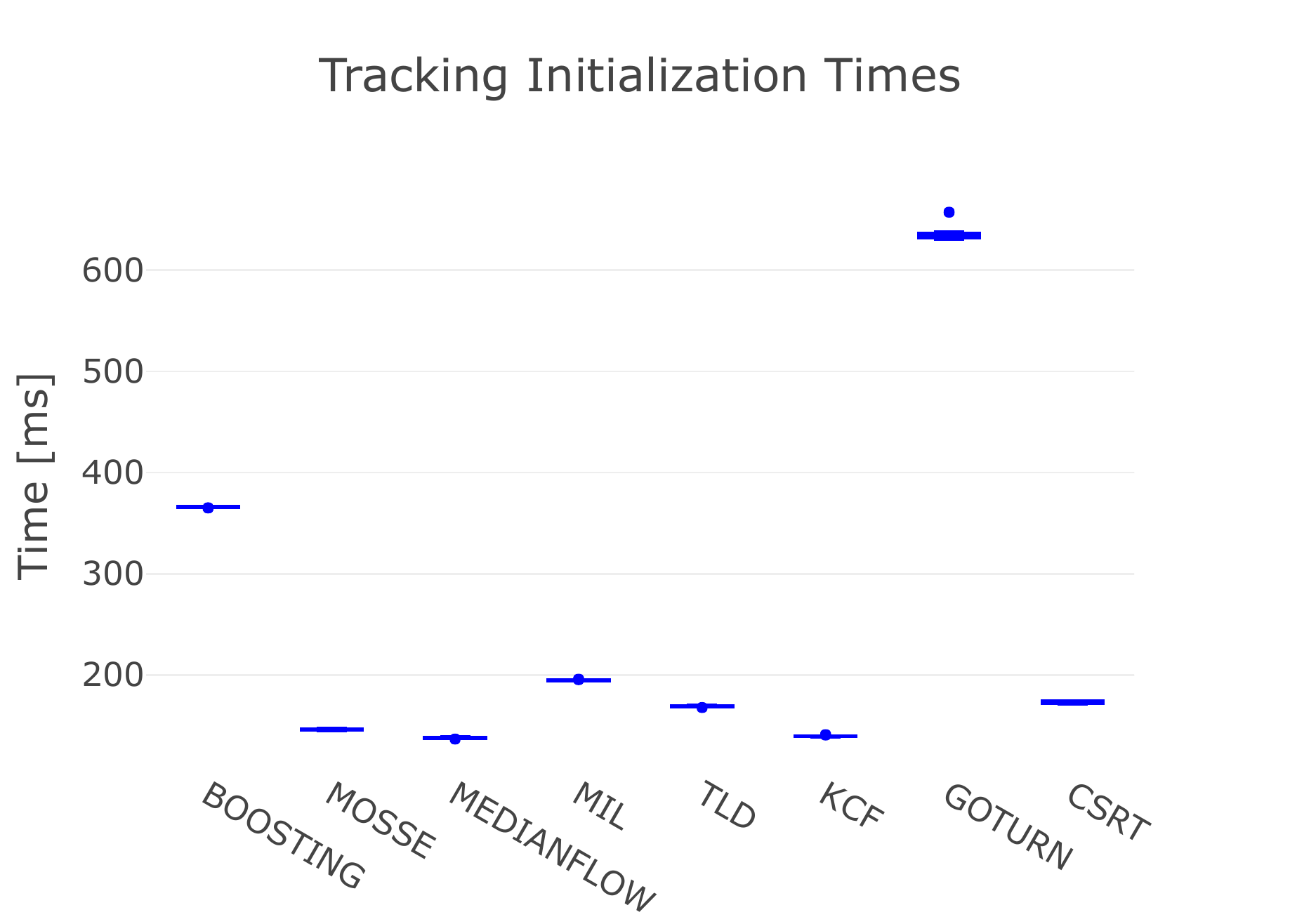}
		\caption{Distribution of the tracker initialization times.} 
		\label{fig:boxplotInitTime}
	\end{center}
\end{figure}

\begin{figure}[]
	\begin{center}
		\includegraphics[width=0.7\textwidth]{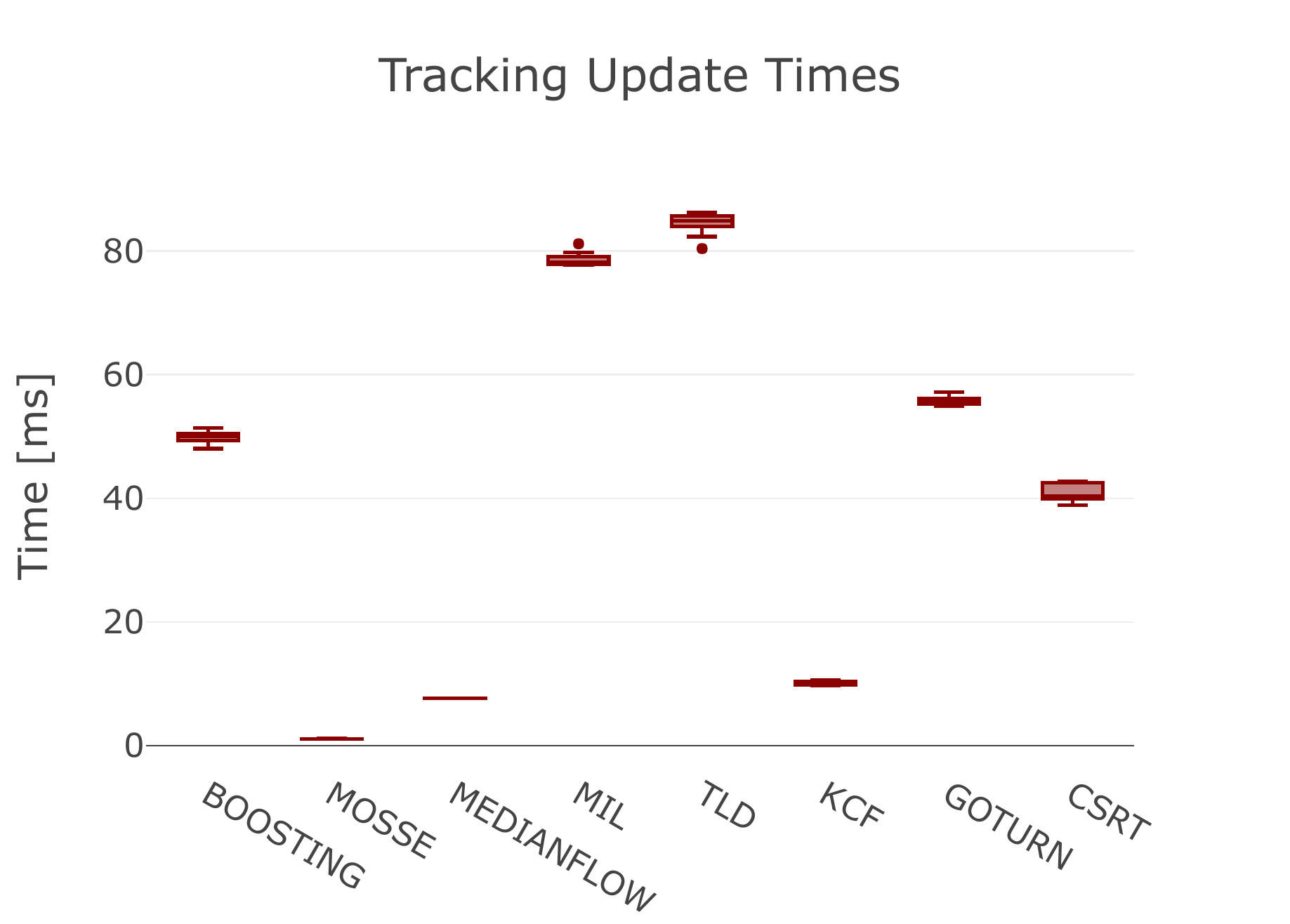}
		\caption{Distribution of the tracker update times.}
		\label{fig:boxplotUpdateTime}
	\end{center}
\end{figure}

\begin{figure}[]
	\begin{center}
		\includegraphics[width=0.9\textwidth]{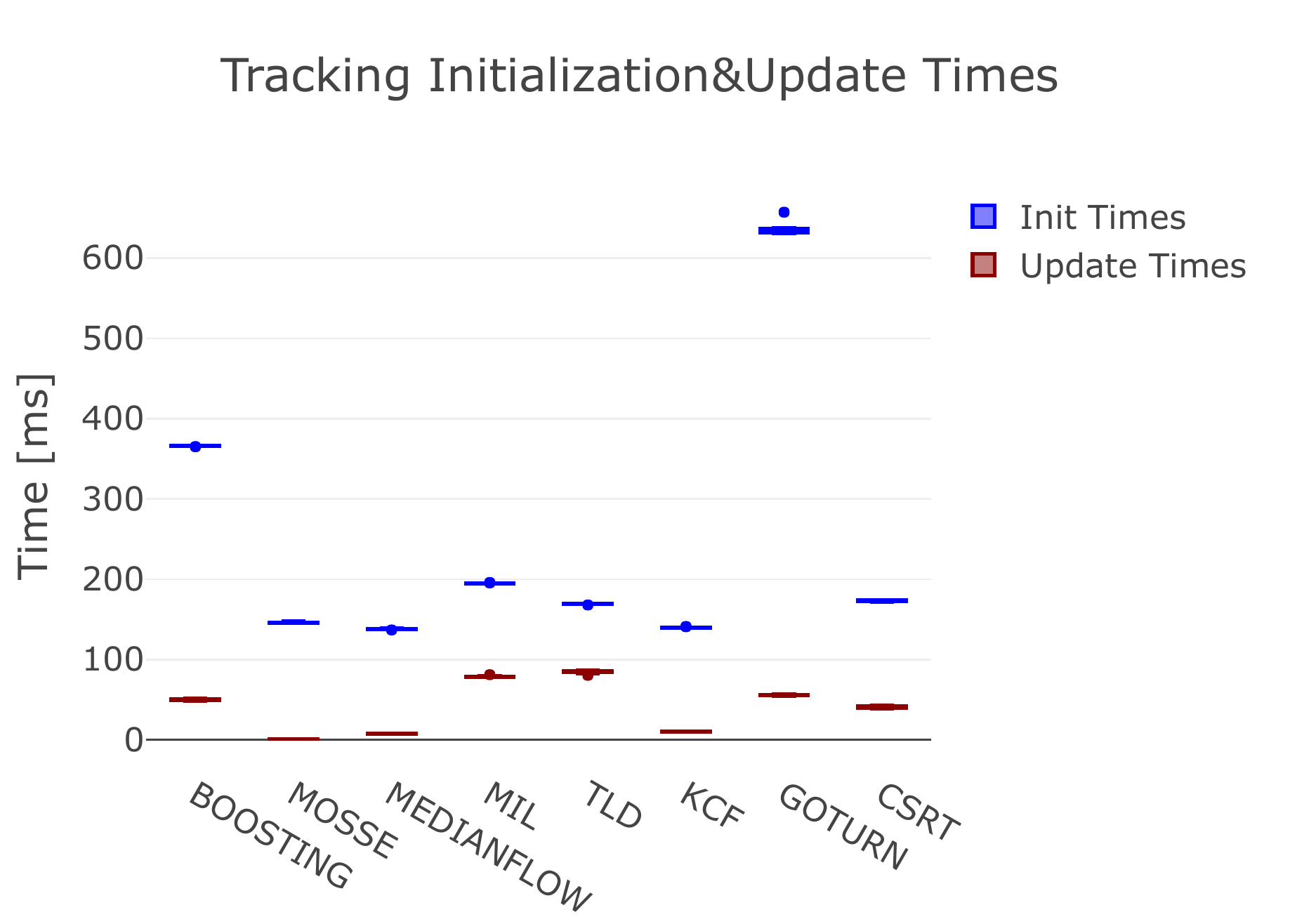}
		\caption{Distribution of the tracker initialization and update times.}
		\label{fig:boxplotTime}
	\end{center}
\end{figure}

\begin{figure}[]
	\begin{center}
		\includegraphics[width=0.8\textwidth]{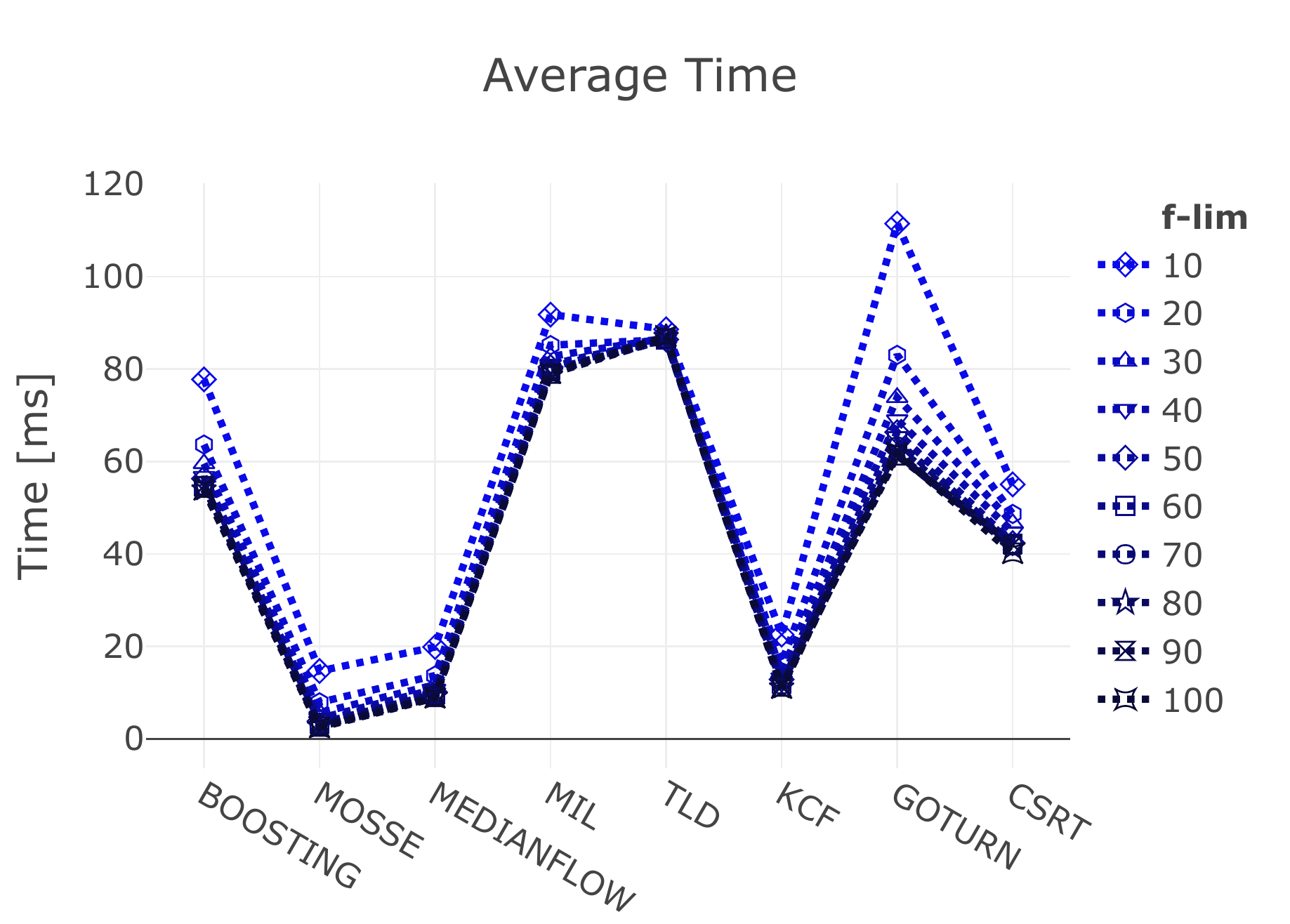}
		\caption{The average running times of trackers.}
		\label{fig:AvgTime}
	\end{center}
\end{figure}

Computational time experiments are performed on a computer with an Intel$^{\circledR}$ Core$^{TM}$ i7-4700MQ CPU clocked at 2.4~GHz and 16~GB memory, running Microsoft$^{\circledR}$ Windows$^{\circledR}$ 8.1 Pro using Python 3.6.2.    The results of the evaluations are delineated in Figure~\ref{fig:boxplotInitTime},~\ref{fig:boxplotUpdateTime},~\ref{fig:boxplotTime} and~\ref{fig:AvgTime}, respectively for tracking initialization times, tracking update times, combined representation of the initialization and update times, and average time demands of the methods for different values of \textit{f-lim}.

BOOSTING and GOTURN require more time to be initialized. Other trackers are initialized in less than 200 ms. In terms of update times, MOSSE, MEDIANFLOW and KCF are the best three trackers followed by CSRT. When average times are considered, MOSSE, MEDIANFLOW, KCF and CSRT again resulting in best four results in order. 

When both tracking accuracy and computational time evaluations are interpreted, MOSSE, KCF and CSRT are the leading three trackers (without any ordering). However, no single tracker overcome the others in all of the tests. While CSRT is prominent in terms of accuracy, MOSSE and KCF are remarkable with less computational time requirement. In any case, CSRT is able to bring down average time requirement under 60 ms which is the reported running time of the detector~\cite{Gokce15}.

\section{Conclusions}\label{conclusions}

In this study, we present a comparative evaluation of various visual tracking methods on mUAV tracking problem by employing them inside a track-by-detection framework. We exploited a detection only method based on cascaded boosted classifiers utilizing local binary bit patterns as feature descriptor, and proposed in~\cite{Gokce15}. We evaluated the tracking methods in terms of tracking perfomance and computational time requirements on outdoor videos. 

Our evaluations revealed that tracking integration to the detection only method enhances the accuracy of locating the mUAV in frames with also an improvement of computational time. None of the trackers is able to beat the others in terms of all aspects. However, MOSSE, KCF and CSRT are noticeable with CSRT is better in terms of accuracy, and the other two are requiring less time. We should indicate that all of these three trackers are based on correlation filters. Selection of the tracker should be done by considering accuracy and available computational power.

\bibliographystyle{splncs04}
\bibliography{references}

\begin{thebibliography}{10}
\providecommand{\url}[1]{\texttt{#1}}
\providecommand{\urlprefix}{URL }
\providecommand{\doi}[1]{https://doi.org/#1}

\bibitem{Aker2018}
Aker, C.: End-To-End Networks For Detection And Tracking Of Micro Unmanned
  Aerial Vehicles. Ph.D. thesis, Middle east Technical University, Ankara,
  Turkey (2018)

\bibitem{andreopoulos201350}
Andreopoulos, A., Tsotsos, J.K.: 50 years of object recognition: Directions
  forward. Computer Vision and Image Understanding  \textbf{117}(8),  827--891
  (2013)

\bibitem{Babenko11}
Babenko, B., Yang, M.H., Belongie, S.J.: Robust object tracking with online
  multiple instance learning. IEEE Trans. Pattern Anal. Mach. Intell.
  \textbf{33}(8),  1619--1632 (2011)

\bibitem{epflAudio}
Basiri, M., Schill, F., Floreano, D., Lima, P.: Audio-based {R}elative
  {P}ositioning {S}ystem for {M}ultiple {M}icro {A}ir {V}ehicle {S}ystems. In:
  Robotics: {S}cience and {S}ystems (RSS) (2013)

\bibitem{Bolme10}
Bolme, D.S., Beveridge, J.R., Draper, B.A., Lui, Y.M.: Visual object tracking
  using adaptive correlation filters. In: CVPR. pp. 2544--2550. IEEE Computer
  Society (2010)

\bibitem{opencv}
Bradski, G.: Opencv. Dr. Dobb's Journal of Software Tools  (2000)

\bibitem{Brewer14}
Brewer, E., Haentjens, G., Gavrilets, V., McGraw, G.: A low swap implementation
  of high integrity relative navigation for small uas. In: Position, Location
  and Navigation Symposium - PLANS 2014, 2014 IEEE/ION. pp. 1183--1187 (May
  2014)

\bibitem{etter}
Etter, W., Martin, P., Mangharam, R.: Cooperative flight guidance of autonomous
  unmanned aerial vehicles. In: CPS Week Workshop on Networks of Cooperating
  Objects (CONET). CPS Week 2011, Chicago (2011)

\bibitem{Gokce15}
Gökçe, F., Üçoluk, G., Şahin, E., Kalkan, S.: Vision-based detection and
  distance estimation of micro unmanned aerial vehicles. Sensors
  \textbf{15}(9),  23805--23846 (2015). \doi{10.3390/s150923805}

\bibitem{Grabner06}
Grabner, H., Grabner, M., Bischof, H.: Real-time tracking via on-line boosting.
  In: Chantler, M.J., Fisher, R.B., Trucco, E. (eds.) BMVC. pp. 47--56. British
  Machine Vision Association (2006)

\bibitem{Held16}
Held, D., Thrun, S., Savarese, S.: Learning to track at 100 fps with deep
  regression networks. In: Leibe, B., Matas, J., Sebe, N., Welling, M. (eds.)
  ECCV (1). Lecture Notes in Computer Science, vol.~9905, pp. 749--765.
  Springer (2016)

\bibitem{Henriques15}
Henriques, J.F., Caseiro, R., Martins, P., Batista, J.: High-speed tracking
  with kernelized correlation filters. IEEE Trans. Pattern Anal. Mach. Intell.
  \textbf{37}(3),  583--596 (2015)

\bibitem{Jaccard}
Jaccard, P.: The distribution of the flora in the {A}lpine zone. New
  Phytologist  \textbf{11}(2),  37--50 (1912).
  \doi{10.1111/j.1469-8137.1912.tb05611.x}

\bibitem{Kalal10}
Kalal, Z., Mikolajczyk, K., Matas, J.: Forward-backward error: Automatic
  detection of tracking failures. In: ICPR. pp. 2756--2759. IEEE Computer
  Society (2010)

\bibitem{Kalal12}
Kalal, Z., Mikolajczyk, K., Matas, J.: Tracking-learning-detection. IEEE Trans.
  Pattern Anal. Mach. Intell.  \textbf{34}(7),  1409--1422 (2012)

\bibitem{Lai2011}
Lai, J., Mejias, L., Ford, J.J.: Airborne vision-based collision-detection
  system. Journal of Field Robotics  \textbf{28}(2),  137--157 (2011).
  \doi{10.1002/rob.20359}

\bibitem{Lin2014}
Lin, F., Peng, K., Dong, X., Zhao, S., Chen, B.M.: Vision-based formation for
  uavs. In: IEEE International Conference on Control Automation (ICCA). pp.
  1375--1380 (June 2014). \doi{10.1109/ICCA.2014.6871124}

\bibitem{Lukezic18}
Lukezic, A., Vojír, T., Zajc, L.C., Matas, J., Kristan, M.: Discriminative
  correlation filter tracker with channel and spatial reliability.
  International Journal of Computer Vision  \textbf{126}(7),  671--688 (2018)

\bibitem{Luo2018}
Luo, H., Xie, W., Wang, X., Zeng, W.: Detect or track: Towards cost-effective
  video object detection/tracking. CoRR  \textbf{abs/1811.05340} (2018),
  \url{http://arxiv.org/abs/1811.05340}

\bibitem{maxim08}
Maxim, P.M., Hettiarachchi, S., Spears, W.M., Spears, D.F., Hamann, J., Kunkel,
  T., Speiser, C.: Trilateration localization for multi-robot teams. In:
  Proceedings of the Sixth International Conference on Informatics in Control,
  Automation and Robotics, Special Session on MultiAgent Robotic Systems
  (ICINCO) (2008)

\bibitem{Moses2011}
Moses, A., Rutherford, M., Valavanis, K.: Radar-based detection and
  identification for miniature air vehicles. In: Control Applications (CCA),
  2011 IEEE International Conference on. pp. 933--940 (Sept 2011).
  \doi{10.1109/CCA.2011.6044363}

\bibitem{Moses2014}
Moses, A., Rutherford, M.J., Kontitsis, M., Valavanis, K.P.: Uav-borne x-band
  radar for collision avoidance. Robotica  \textbf{32},  97--114 (1 2014).
  \doi{10.1017/S0263574713000659}

\bibitem{Nishitani06}
Nishitani, A., Nishida, Y., Mizoguch, H.: Omnidirectional ultrasonic location
  sensor. In: IEEE Conference on Sensors (2005)

\bibitem{Opromolla18}
Opromolla, R., Fasano, G., Accardo, D.: A vision-based approach to uav
  detection and tracking in cooperative applications. Sensors  \textbf{18}(10)
  (2018). \doi{10.3390/s18103391}

\bibitem{Raharijaona2015}
Raharijaona, T., Mignon, P., Juston, R., Kerhuel, L., Viollet, S.: Hypercube: A
  small lensless position sensing device for the tracking of flickering
  infrared leds. Sensors  \textbf{15}(7),  16484 (2015).
  \doi{10.3390/s150716484}

\bibitem{rivard08}
Rivard, F., Bisson, J., Michaud, F., Letourneau, D.: Ultrasonic relative
  positioning for multi-robot systems. In: IEEE International Conference on
  Robotics and Automation (ICRA). pp. 323 --328 (may 2008).
  \doi{10.1109/ROBOT.2008.4543228}

\bibitem{roberts2}
Roberts, J., Stirling, T., Zufferey, J., Floreano, D.: 3-d relative positioning
  sensor for indoor flying robots. Autonomous Robots  \textbf{33}(1-2),  5--20
  (2012)

\bibitem{roberts1}
Roberts, J.: Enabling {C}ollective {O}peration of {I}ndoor {F}lying {R}obots.
  Ph.D. thesis, Ecole Polytechnique Federale de Lausanne (EPFL) (2011),
  \url{http://library.epfl.ch/theses/?nr=5078}

\bibitem{stirling}
Stirling, T., Roberts, J., Zufferey, J., Floreano, D.: Indoor navigation with a
  swarm of flying robots. IEEE International Conference on Robotics and
  Automation (ICRA)  (2012)

\bibitem{Tijs2010}
Tijs, E., de~Croon, G., Wind, J., Remes, B., de~Wagter, C., de~Bree, H.E.,
  Ruijsink, R.: Hear-and-avoid for micro air vehicless. In: International Micro
  Air Vehicle Conference and Competitions (IMAV) (2010)

\bibitem{vasarhelyi2014outdoor}
V{\'a}s{\'a}rhelyi, G., Vir{\'a}gh, C., Somorjai, G., Tarcai, N., Szorenyi, T.,
  Nepusz, T., Vicsek, T.: Outdoor flocking and formation flight with autonomous
  aerial robots. In: IEEE/RSJ International Conference on Intelligent Robots
  and Systems (IROS). pp. 3866--3873 (2014)

\bibitem{Welsby01}
Welsby, J., Melhuish, C., Lane, C., Qy, B.: Autonomous minimalist following in
  three dimensions: A study with small-scale dirigibles. In: Proceedings of
  Towards Intelligent Mobile Robots (2001)

\bibitem{Zhang2014}
Zhang, M., Lin, F., Chen, B.M.: Vision-based detection and pose estimation for
  formation of micro aerial vehicles. In: International Conference on
  Automation Robotics Vision (ICARCV). pp. 1473--1478 (Dec 2014).
  \doi{10.1109/ICARCV.2014.7064533}

\end{thebibliography}

\end{document}